\documentclass[fullpaper]{nldl}

\renewcommand{\DOI}{12345} 

\usepackage[utf8]{inputenc}
\usepackage{url}
\usepackage{graphicx}
\usepackage{authblk}
\usepackage{subcaption}
\usepackage{tabularx}

\usepackage{algorithm}
\usepackage{algorithmic}

\usepackage[square,numbers]{natbib}

\usepackage{hyperref}

\usepackage{multirow}
\usepackage{amsmath}

\title{Lidar-based Norwegian tree species detection using deep learning}

\author[1]{Martijn Vermeer}  
\author[1]{Jacob Alexander Hay\thanks{Corresponding Author: hay@stcorp.no}} 
\author[1]{David Völgyes} 
\author[2]{Zsófia Koma} 
\author[2]{Johannes~Breidenbach} 

\author[1]{Daniele~Stefano~Maria~Fantin} 

\affil[1]{Science and Technology AS}
\affil[2]{Norwegian Institute of Bioeconomy Research}

\date{\vspace{-5ex}}

\newcommand{\Fone}{$\mathrm{F}_1$}

\begin{document}

\nldlmaketitle
\begin{abstract}
	{\bf Background}:
	The mapping of tree species within Norwegian forests is a time-consuming process,
	involving forest associations relying on manual labeling by experts. The process
	can involve both aerial imagery, personal familiarity, or on-scene references,
	and remote sensing data. The state-of-the-art methods usually use high
	resolution aerial imagery with semantic segmentation methods.\newline
	{\bf Methods}: We present a deep learning based tree species classification model
	utilizing only lidar (Light Detection And Ranging) data.
	The lidar images are segmented into four classes (Norway Spruce, Scots Pine, Birch, background)
	with a U-Net based network. The model is trained with focal loss over
	partial weak labels. A major benefit of the approach is that both the lidar imagery and
	the base map for the labels have free and open access.\newline
	{\bf Results}: Our tree species classification model achieves a macro-averaged \Fone score
	of 0.70 on an independent validation with National Forest Inventory (NFI) in-situ sample plots.
	That is close to, but below the performance of aerial, or aerial and lidar combined models.
\end{abstract}

\section{Introduction}
\subsection{Motivation}
\label{ssec:motivation}
Forest management and biodiversity hotspot identification require detailed,
spatially continuous mapping of forests and tree species~\cite{PECCHI2019108817, felton2020tree}.
These maps are ideally up-to-date and complete.
Norway has had a National Forest Inventory since the 1920s~\cite{breidenbach2021national},
and model-assisted national mapping (SR16) is available~\cite{sr16}, but often higher
resolution maps are required for forest management tasks.

In Norway, the three main species used for production are Norway spruce, Scots pine,
and Birch (more generally: broadleaf species).
Current methods for mapping the spatial distribution rely heavily on updating previous maps
using aerial and lidar sources.
This multivariate data is processed by forestry experts, who carry out manual classification.
Forest inventories are therefore both time-consuming and expensive.
For this reason, maps are often reused between updates, and only
actively logged stands will be passed through expert validation, which introduces a
possible drift between the map and the actual stands.

To tackle these challenges, deep learning-based semantic segmentation seemed to
be an obvious choice. Indeed, semantic segmentation with deep learning has seen increased usage within
forest monitoring~\cite{KATTENBORN202124,nezami2020tree}.

We only found studies that are using additional data sources besides lidar, or use higher
resolution gridded or point cloud data.
Therefore, we performed a feasibility study to establish whether models trained on 1~m
resolution, gridded lidar data is comparable to 0.2~m resolution RGB images~\cite{norgeibilder} for Norwegian
tree species mapping.

\section{Background and related work}
\label{ssec:background}
Machine learning methods used for tree species identification have recently shifted
from model-assisted methods~\cite{sr16} and traditional machine learning-based
methods~\cite{rs4092661,FASSNACHT201664}
towards deep learning~\cite{KATTENBORN202124,nezami2020tree}.
Deep learning is able to utilize contextual and texture information,
which opens up opportunities for better utilization of the information available
in the very high-resolution imagery data and open lidar data products.

Lidar frequently complements other data sources~\cite{sr16,GHOSH201449}
as the canopy height and tree-crown structure are a strong predictor variable for tree species identification.
Other deep learning approaches utilize the point cloud itself for classifying tree species.
However, these are rarely applicable across large areas due to the need for intensive data processing
and often aim for single-tree classification~\cite{Mustafic2022DeepLF,Liu2022TreeSC}.
Overall, related published deep learning models
are not directly comparable to a model that uses rasterized medium-resolution lidar-derived data products
such as Digital Surface and Digital Terrain models.
The importance of resolution is highlighted in~\cite{Gan2023TreeCD}, among others.

\section{Materials and Methods}

\subsection{Data sources}

We used three main data sources for training the segmentation model:
\begin{enumerate}
	\item Digital terrain model (DTM)\footnote{\url{https://www.kartverket.no/geodataarbeid/nasjonal-detaljert-hoydemodell}} with a resolution of 1~m.
	\item Digital surface model (DSM) with a resolution of 1~m.
	\item Norwegian forest resources map (Skogressurskart, SR16~\cite{sr16}) with a resolution of 16~m.
\end{enumerate}
Input features were derived from the DTM and DSM, and SR16 served as a weak label.

In addition, we used land borders from OpenStreetMap~\cite{OpenStreetMap} to limit the area
only to land and exclude sea and major lakes.
For internal validation, we split the data into train and validation regions.

The National Forest Inventory (NFI) was used as test data.
The NFI is a continuous inventory system of permanent plots, with $\frac{1}{5}$ of the plots are measured every year.
The permanent sample grid covers all land use classes, and if it does not contain trees,
the land use class is assigned based on visual interpretation of aerial images.
In forests, more than 120 variables are measured in circular plots of 250~$\mathrm{m}^2$.
For instance, the timber volume of each main tree species (Norway spruce, Scots pine, and Birch) with diameter $>5$~cm has been recorded.
More details can be found in~\cite{Breidenbach2020ACO}.

In this study, we used the plots measured in the 2017-2021 NFI cycle within the 3~km~$\times$~3~km grid.
Split plots had been excluded from the analysis.
The plots without stocked volume were set as the "background" class, and the dominant
tree species classes were determined based on the timber volume.

\subsection{Lidar pre-processing}

The lidar data was downloaded at a resolution of 1~m in gridded surface and elevation models.
To capture tree canopies, we generated what is known as a canopy
height model, which is the difference between the surface and elevation models:
$$
	\mathrm{CHM} = \mathrm{DSM} - \mathrm{DTM}
$$

\subsection{Label pre-processing}
There are several challenges with matching 1~m resolution lidar data
with 16~m resolution label data (SR16).

The first challenge is the time-wise mismatch.
Both the SR16 map creation and the lidar data collection campaign took several years.
This led to local mismatches where either canopy loss is not yet visible on
SR16 but present in the lidar data, e.g. due to construction. The opposite
also appeared in the data: SR16 missing areas that are clearly forested in the lidar data.
This could be due to a later canopy loss that happened after the lidar campaign but before
the SR16 update.

The second problem is the resolution of the data: 16~m label pixels which
have a forest border approximately in the center of the pixels will inevitably
contain almost 50\% error by rounding the pixel into one of the classes.
When such label is resampled to 1~m resolution, it could yield a several meters wide
band of mislabeled pixels at the forest border due to the coarse resolution of the original label.

Finally, the SR16 map only contains tree species within forested areas,
while at 1~m resolution, canopy openings would fall into the "background" class that does not exist in SR16 within forests.

In order to address these issues, we used the following procedure to pre-process the map into training labels.
First, all unlabeled pixels were set to the background class.
Second, all forest/no-forest border pixels were set to unlabeled, effectively
tackling the forest borders.
After this step, the forest map was upsampled to 1~m resolution
by nearest neighbor upsampling.
Finally, an 11~m~$\times$~11~m median filter was applied to the CHM data. Every location where the
CHM median value was below 0.3~m was labelled as background.

A second round of labeling was performed after a full training. In this second round,
larger areas (100 SR16 pixels (25600~$\mathrm{m}^2$) or larger) which were consistently labeled as background and predicted as any kind of forest,
or forest label that was consistently predicted as background, got unlabeled.
That step is an attempt to remove systematic labeling errors which could severely
impact classification performance~\cite{Maiti2022EFFECTOL}
when the label and the features are different due to e.g. recent clearcuts.
This attempt to reduce label noise reduced the total labeled area insignificantly
($<0.1\%$).
A second training was performed with this refined, partially unlabeled data,
and only this second training on clean data is reported as result.

\subsection{Deep Learning model}
\enlargethispage{3mm}

A standard U-Net network~\cite{ronneberger2015u} was used.
The network parameters are summarized in Table~\ref{table:hp}.

\begin{table}
	\centering
	\caption{Summary Parameters of a U-Net Network}
	\label{table:hp}
	\begin{tabularx}{\linewidth}{c|X}
		\textbf{Parameter}        & \textbf{Value}        \\
		\hline
		Image size                & 2048 $\times$ 2048 px \\
		Input channels            & 2 (DTM, CHM)          \\
		Output channels           & 4                     \\
		Depth                     & 5                     \\
		Initial number of filters & 16                    \\
		Upsampling method         & convolution           \\
		Padding method            & reflection            \\
		Normalization             & instance              \\
	\end{tabularx}
\end{table}

At inference time, the following extra post-processing steps were applied:
\begin{enumerate}
	\item The prediction logits were blurred with a Gaussian filter ($\sigma=1$~px)
	      in order to reduce single pixel noise.
	\item The tiles were cropped by removing a 64 pixel wide edge
	      in order to reduce edge artifacts originating from tiling.
\end{enumerate}

\subsection{Loss function}

Focal loss~\cite{lin2017focal} with $\gamma=3$ was selected in order
to reduce the contribution of easy samples. The softmaxed logits
were cut off at a $p\leq 0.1$ level in order to avoid contribution from mislabeled pixels.
The classes were weighted inversely with their frequency.

\subsection{Augmentation}

As augmentation of the raw data, standard geometric transformations were used (i.e. rotation
and flipping). We did not change the brightness as the lidar values have a direct impact on
the interpretation of species.

\begin{figure*}[htb!] 
	\centering{}

	\begin{subfigure}{0.3\textwidth}
		\includegraphics[width=\linewidth]{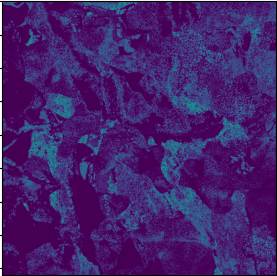}
		\caption{canopy height model}
	\end{subfigure}
	\hfill
	\begin{subfigure}{0.3\textwidth}
		\includegraphics[width=\linewidth]{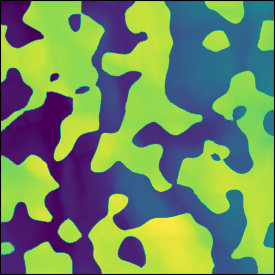}
		\caption{digital terrain model}
	\end{subfigure}
	\hfill
	\begin{subfigure}{0.3\textwidth}
		\includegraphics[width=\linewidth]{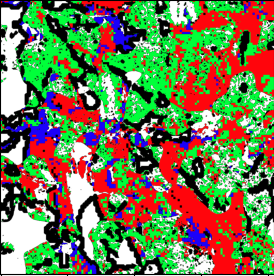}
		\caption{species labels and background}
	\end{subfigure}

	\caption{Cowmix augmentation: mixing two images in cow patterns introduces, best visible on the elevation
		data (B), but also noticeable in the two other subfigures. Species are in RGB, background is white, unlabeled is black.}
	\label{fig:cowmix}
\end{figure*}

Another less known augmentation was used, which we call 'CowBatchMix' (see Fig.~\ref{fig:cowmix}).
Originating from the 'CowMix' augmentation~\cite{cowmix}, this augmentation mixes two batch samples
in a cow pattern. The reasoning for its selection was that the forest edges were unlabeled,
therefore, the network might have difficulties learning forest -- no-forest transitions
due to the lack of sharp transitions. Close to edge pixels are hard to classify even
in fully supervised settings which could be mitigated by
distance maps-based positional weighting~\cite{ronneberger2015u} when borders are known,
or by downweighting already learned samples, see our focal loss~\cite{lin2017focal} choice.
By mixing the images with CowMix, artificial edges were introduced with sharp transitions.
This incentivizes the network to both learn local features and be precise about transitions.
Furthermore, it acts as a regularizer against overfitting.

\subsection{Study area}

\begin{figure}
	\centering
	\includegraphics[width=0.5\textwidth]{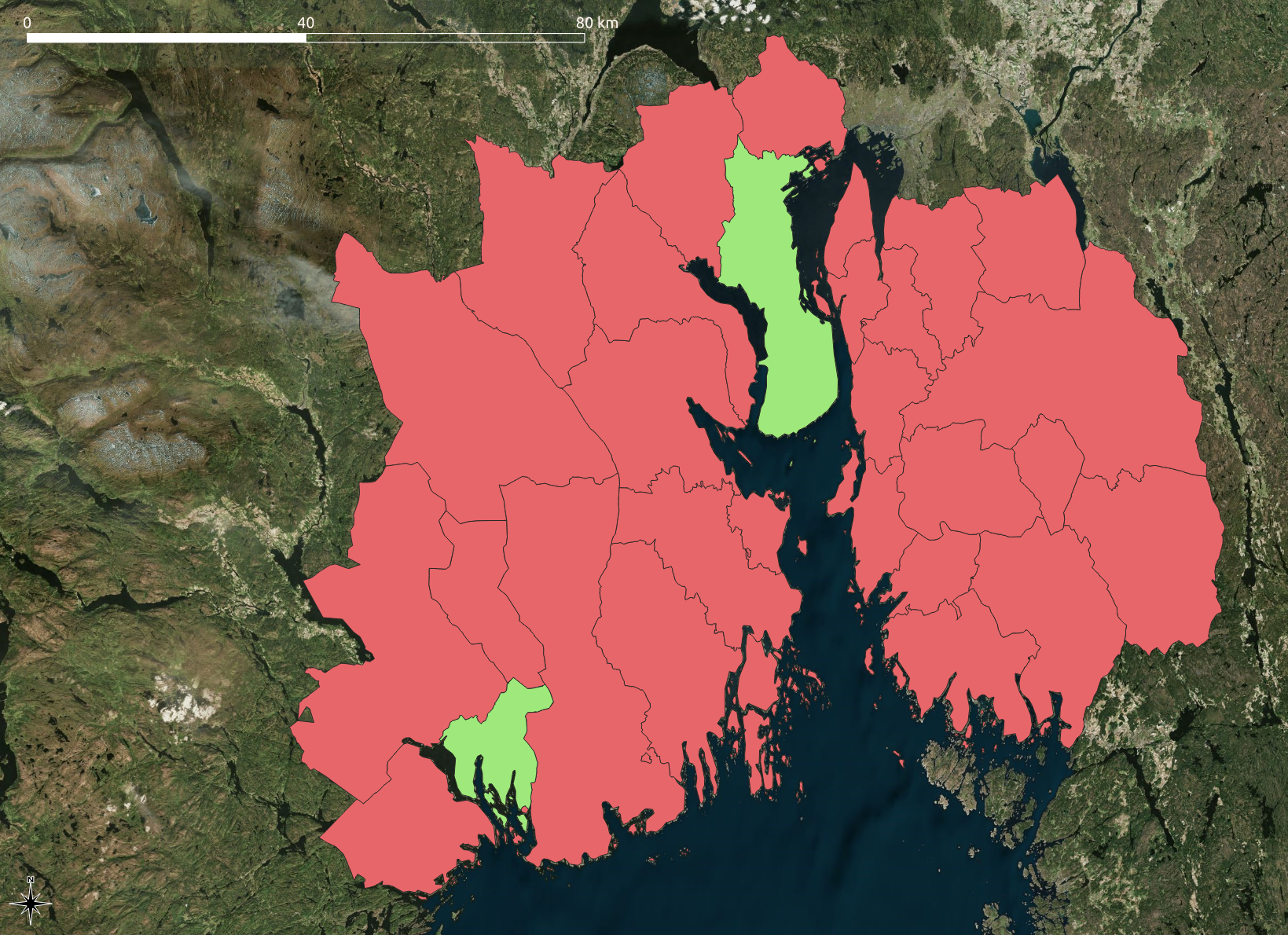}
	\caption{Training (red) and validation (green) data split in the study area
		(Viken, Norway).}
	\label{fig:data_split}
\end{figure}

The study area was a large area south of Oslo, Norway.
The data was split into train (8869~$\mathrm{km}^2$) and validation data (541~$\mathrm{km}^2$), depicted in Fig.~\ref{fig:data_split}.
The independent test data was the NFI sample plots. The exact locations are not public.

\section{Results}
The full study area was evaluated using the independent NFI plots.
If the plots were mixed, either in the reference or in the prediction,
the largest contributor was used as the 'dominant species' or background class.

An inherent limitation of the evaluation comes from the plots.
If the plot contains a mix of species,
then small prediction errors might be amplified.
For instance, a prediction might mismatch by just a few percent, but e.g., 45\%-55\%
reference and 55\%-45\% for prediction is accounted as a single miss,
just like a 0\%-100\% vs 100\%-0\% mismatch.

\begin{figure*}[htb!] 
	\centering
	\begin{subfigure}{0.45\textwidth}
		\includegraphics[width=\linewidth]{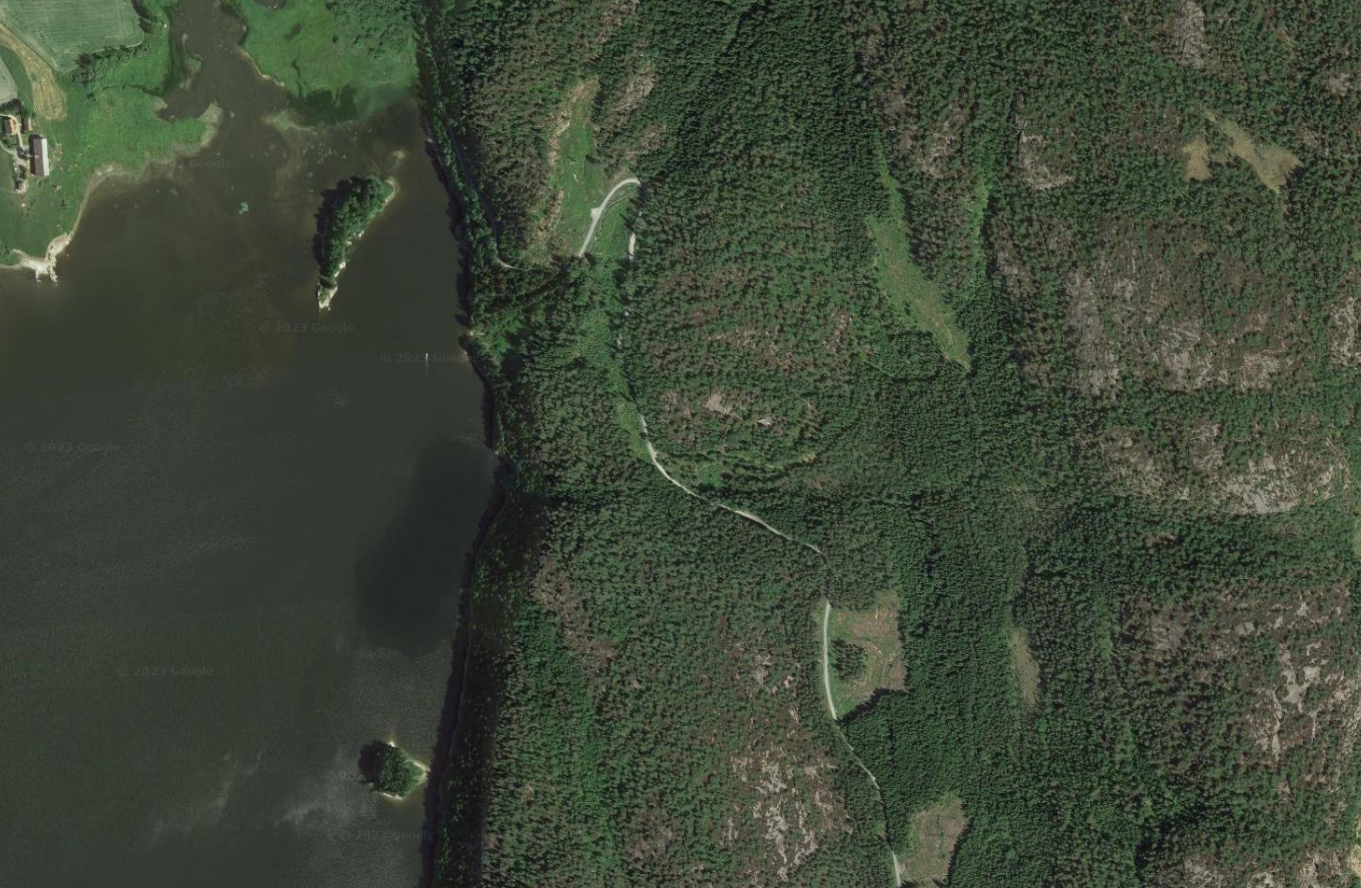}
		\caption{Google Maps satellite view}
	\end{subfigure}
	\hfill
	\begin{subfigure}{0.45\textwidth}
		\includegraphics[width=\linewidth]{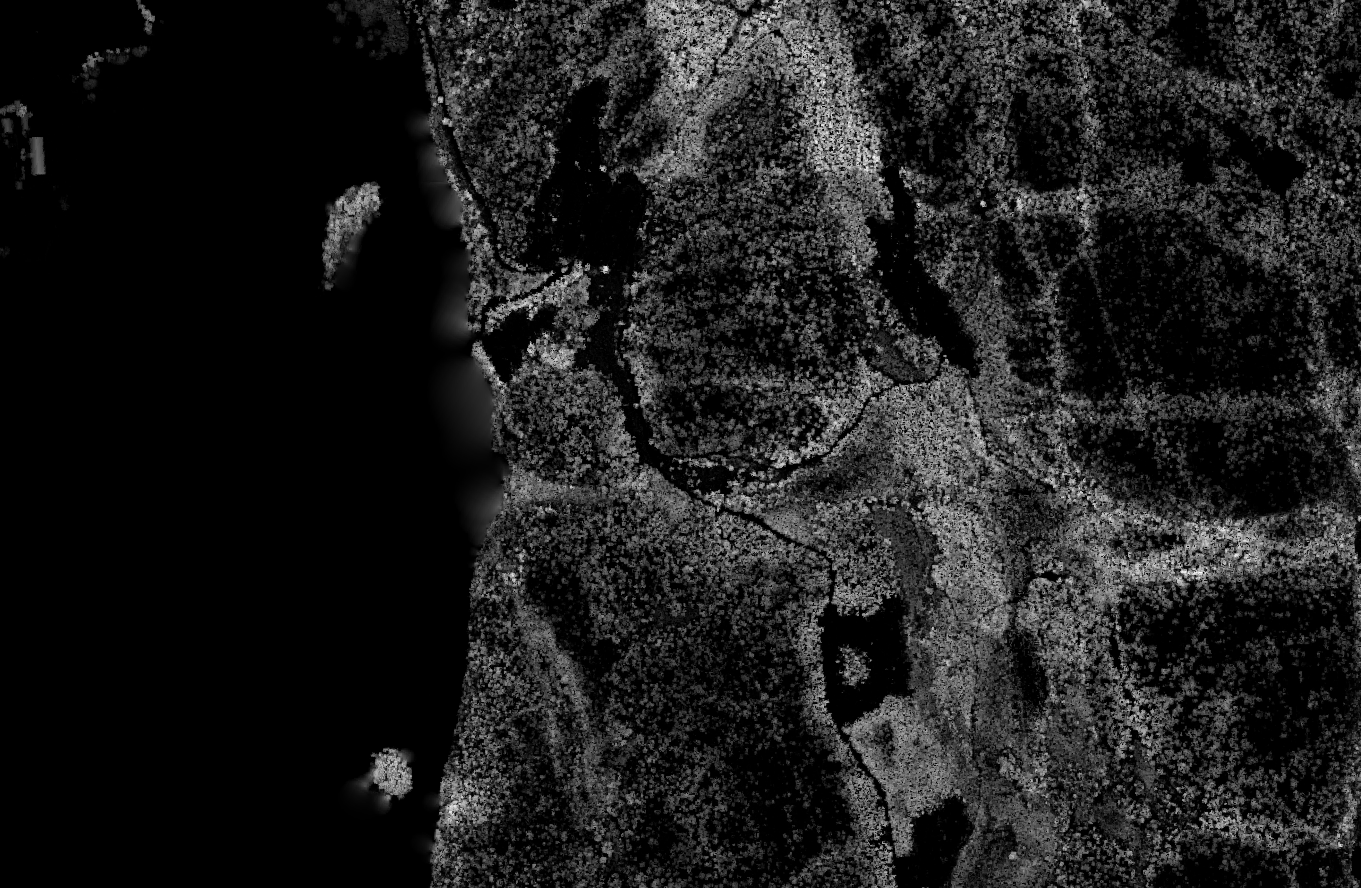}
		\caption{Canopy Height Model}
	\end{subfigure}
	\hfill

	\begin{subfigure}{0.45\textwidth}
		\includegraphics[width=\linewidth]{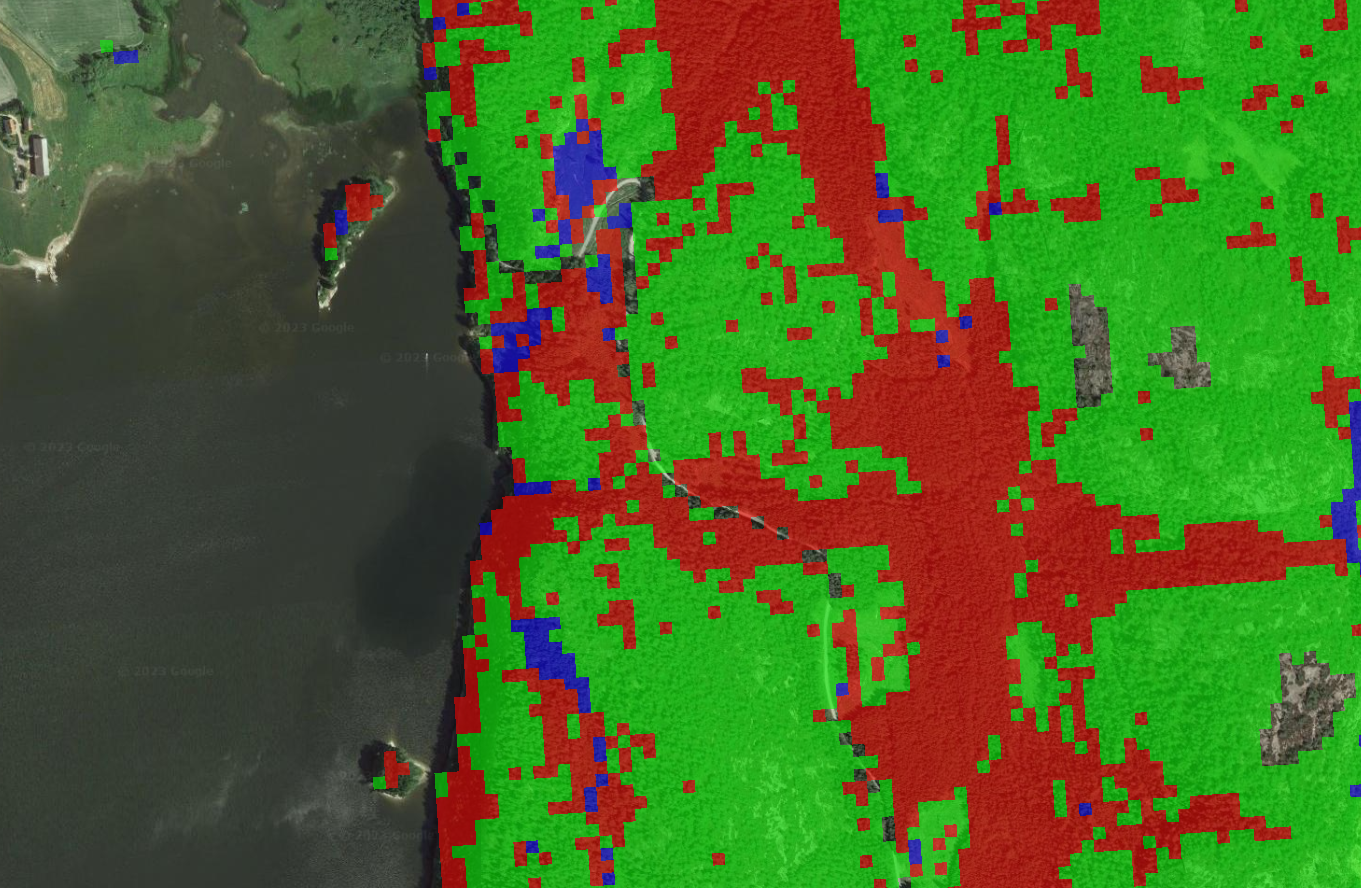}
		\caption{Satellite view overlaid with unmodified SR16}
	\end{subfigure}
	\hfill
	\begin{subfigure}{0.45\textwidth}
		\includegraphics[width=\linewidth]{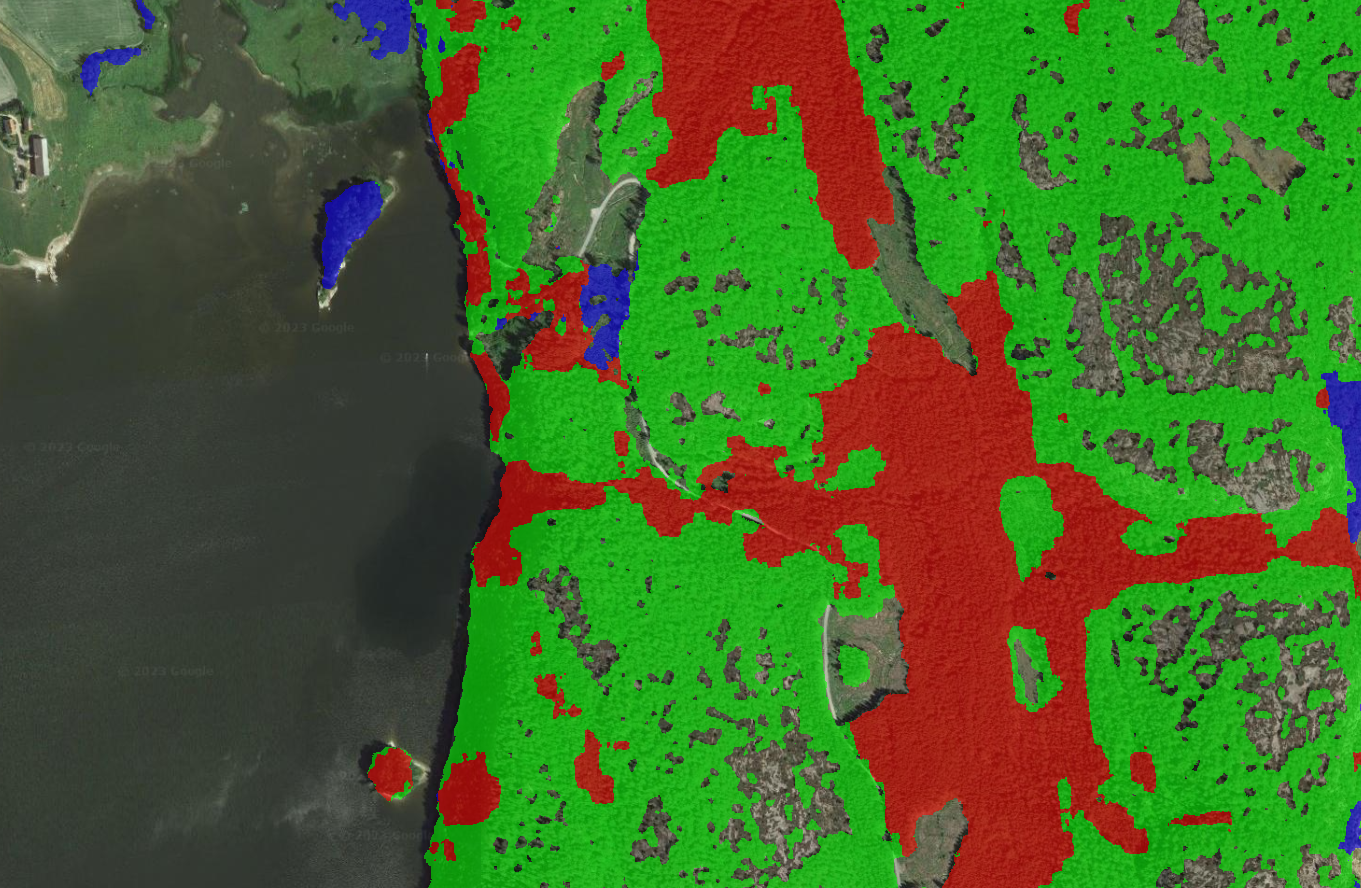}
		\caption{Overlay with our model's prediction}
	\end{subfigure}
	\hfill
	\caption{Example prediction: the high-resolution prediction follows local edges more closely.}
	\label{fig:example}
\end{figure*}

The confusion matrix results with derived metrics (precision, recall, \Fone score)
are presented in Table~\ref{table:result}.
In addition, an example region is visualized in Fig.~\ref{fig:example}
to give a qualitative impression.

\begin{table*}[htp!]
	\centering
	\vspace{3mm}
	\caption{Testing results: confusion matrix, per-class recall, precision, \Fone score,
		overall accuracy (OA), and macro-averaged \Fone score.}
	\label{table:result}
	\vspace{-3mm}
	\resizebox{\textwidth}{!}{
		\begin{tabular}{c|c|c c c c|c|c|c|}
			\multicolumn{2}{c}{} & \multicolumn{4}{c}{NFI plots} & \multicolumn{2}{c}{}                                                                                                     \\
			\cline{2-9}
			                     & All                           & Background           & Birch       & Scots Pine   & Norway Spruce & $\sum$               & Precision & \Fone Score       \\
			\cline{2-9}
			\multirow{4}{*}{\rotatebox{90}{Predictions}}
			                     & Background                    & \textbf{352}         & 14          & 23           & 23            & 412                  & 0.85      & 0.90              \\
			                     & Birch                         & 9                    & \textbf{64} & 15           & 36            & 124                  & 0.52      & 0.52              \\
			                     & Scots pine                    & 6                    & 17          & \textbf{153} & 43            & 219                  & 0.70      & 0.69              \\
			                     & Norway spruce                 & 4                    & 26          & 34           & \textbf{187}  & 251                  & 0.75      & 0.69              \\
			\cline{2-9}
			                     & $\sum$                        & 371                  & 121         & 225          & 289           & 1006                 & OA: 0.75  & Macro \Fone: 0.70 \\
			\cline{2-9}
			                     & Recall                        & 0.95                 & 0.53        & 0.68         & 0.65          & \multicolumn{3}{c}{}                                 \\
			\cline{2-6}
		\end{tabular}}
\end{table*}

\section{Discussion}

\subsection{Limitations of the validation}
The training data is as independent of the testing data as possible.
In Norway, NIBIO produces a 16~m resolution forest resource map (SR16)~\cite{sr16}.
This map is model-assisted and generated in an automatic way with traditional machine learning methods.
Since we use the SR16 map as input, there is a slight dependency chain between our input data and the test data.
To the best of our knowledge, SR16 is used in all known Norwegian forest maps,
including private companies' own inventories where they often use SR16 as a base
and update their local area.
We are not aware of any publicly available Norwegian tree species map that does
not rely at least partially on SR16 and therefore indirectly on the NFI plots.
This is an inherent limitation, and we did not see any way to overcome this challenge.
The difference between our model output and SR16 maps shows that they are structurally different.

\subsection{The benefits of the lidar source}
There are several benefits of using lidar only for tree species detection.
First, lidar is an active source; it lacks terrain shadows, which frequently cause problems in satellite or aerial imaging.
Similarly, atmospheric effects do not affect it because campaigns are performed
in clear sky conditions, and elevation measurements use time-of-flight
measurements, which removes the issue with local reflectance (i.e., color of the
forest).
Finally, another significant benefit is the national coverage with freely available data.
Aerial imagery campaigns also exist for Norway~\cite{norgeibilder}, but these datasets are expensive.

\subsection{The drawbacks of the lidar source}
Several drawbacks are also present. First and foremost is the resolution.
The national 1~m gridded data is very likely on the edge of feasibility, and seasonal changes of leaves could affect the detection of deciduous species.
Moreover, the shape of the tree and the shape of the forest are not independent.
In dense forest, the canopy can form a continuous unit where the canopy shape is
the result of the interaction between trees, while in sparse forests the shape
of individual trees can be more easily recognized.
This issue is especially challenging with freshly planted, dense forest where the trees are small and densely packed.
These forest structures are not only dependent on the species themselves and the age of the trees but also on the soil, water access, climate, etc.
Due to these limitations, we think the model is regionally applicable: it is
validated in the same type of area, but if climate, soil, potential species
change, then a new regional model would be necessary.
While the study area was selected with these issues in mind and tried to incorporate diverse data, verification on a larger scale remains future work.

In addition, lidar cannot capture spectral information. Spectral and hyperspectral information
can be very important for tree species classification~\cite{6331005} when the number of species are larger and the species
have similar height and crown structure.
These regional differences in tree species distribution makes it hard to compare
our results to related work. A study~\cite{vermeer2023semi} performed in the middle of Norway
reported macro-averaged \Fone score of approximately 0.75 by using semi-supervised
deep learning on 20cm resolution aerial imagery by using weak labels.
While this study achieved slightly higher score (0.75) than our results (0.70),
it used a significantly higher resolution input. On top of that the region is also slightly different,
and it was compared to forest stand level majority classes instead of pointwise comparison.

\subsection{Comparison to SR16}
We have compared  our lidar-based tree species classification map with the SR16-based tree species map.
We have found that some errors from SR16 are independently distributed in the training data and the only-lidar map does not reproduce the same errors.
This improvement can be the result of using the shape of the trees in training.
However, both methods (SR16 and the one presented in this paper) are sensitive to the density of the forests.
Density-related systematic mislabels in SR16 might also be learned by our model.
On the other hand, our model is more precise with regards to the forest
borders, including but not limited to forest roads and sparse forests with open
spaces.
While these results are not visible in the numerical comparison, we argue that
this increased resolution has great utility for planning (e.g., housing,
construction work), where the exact species are less important than the forest
borders.

\subsection{Impact and utility}
Using a free data source and achieving comparable results to aerial-based tree
species classification has the potential to reduce cost of accurate forest mapping.
Some potential application areas:
\begin{enumerate}
	\item Improving the accuracy of the estimation of forest attributes such as timber volume and biomass.
	\item High resolution mapping of tree species is important for planning sustainable forest management~\cite{PECCHI2019108817}.
	\item It can provide important information for climate adaptation and mitigation and biodiversity assessment
	      and biodiversity hotspot identification~\cite{felton2020tree}.
\end{enumerate}

\section{Conclusion}

Our work suggests that the use of lidar-only methods is not only feasible
with regards to detection and segmentation of Norwegian tree species,
but it also comes with the benefits of utilizing publically available, open access data.
This allows the creation of larger, higher resolution maps,
without dependence upon commercial data, at comparatively lower cost,
without sacrificing much in terms of performance.

\section{Future work}
A current limitation concerns the study area. Although it encompasses approximately
$9400~\mathrm{km}^2$, it is confined to a specific region of Norway.
For our internal test, we produced a country-wide tree species map.
However, the independent validation was solely focused on the South of Norway region, as detailed in earlier sections.
In upcoming work, we anticipate extending the validation from this regional scope to a national one, and possibly to neighboring countries.

A potential direction for future work lies in enhancing prediction accuracy by integrating higher quality data sources.
Examples include additional gridded lidar metrics, higher resolution gridded data, or more refined point cloud data.

Subsequent ablation studies might delve into evaluating the significance of particular choices,
such as the label noise removal technique or the influence of the CowMix augmentation.

\section*{Acknowledgment}
This work was funded by the European Space Agency (ESA) as part of the SENTREE project
(ESA Contract No 4000136015/21/I-DT-lr) in the ``ESA AO/1-10468/20/I-FvO FUTURE
EO-1 EO SCIENCE FOR SOCIETY PERMANENTLY OPEN CALL'' program.

The authors thank Tord Kriznik Sørensen for his contributions to the technical
infrastructure of the pre-cursor project, and the discussions about species detection.

\bibliographystyle{abbrvnat}
\bibliography{references}

\end{document}